# Fuzzy Logic Interpretation of Quadratic Networks

Fenglei Fan, Ge Wang*

Biomedical Imaging Center, BME/CBIS

Rensselaer Polytechnic Institute, Troy, New York, USA

fanf2@rpi.edu, ge-wang@ieee.org

*Abstract* — **Over past several years, deep learning has achieved huge successes in various applications. However, such a data-driven approach is often criticized for lack of interpretability. Recently, we proposed artificial quadratic neural networks consisting of quadratic neurons in potentially many layers. In cellular level, a quadratic function is used to replace the inner product in a traditional neuron, and then undergoes a nonlinear activation. With a single quadratic neuron, any fuzzy logic operation, such as XOR, can be implemented. In this sense, any deep network constructed with quadratic neurons can be interpreted as a deep fuzzy logic system. Since traditional neural networks and quadratic counterparts can represent each other and fuzzy logic operations are naturally implemented in quadratic neural networks, it is plausible to explain how a deep neural network works with a quadratic network as the system model. In this paper, we generalize and categorize fuzzy logic operations implementable with individual quadratic neurons, and then perform statistical/information theoretic analyses of exemplary quadratic neural networks.**

## I. INTRODUCTION

In the field of machine learning, artificial neural networks (ANNs), especially deep neural networks (CNNs), have recently achieved impressive successes in important applications including classification, unsupervised learning, prediction, image processing and analysis, as well as tomographic reconstruction [1-3]. However, a neural network is often considered as a black box without interpretability. Questions are often asked include what a neural network does in generating its output, and more importantly how a network can be designed and optimized. The deep learning researchers definitely want to have the model that is explainable, and modifiable under guidelines. Lacking explainability has become a primary obstacle to the wide-spread adaptation and further development of deep learning techniques [4]. Since understanding the mechanism of deep neural networks is challenging, over recent years extensive efforts have been made along this direction. Heuristically, existing methods can be categorized into three classes based on the categorization of [5]. Here we add two more classes: mathematical/physical models and neural network visualization, which give unique insights from different perspectives. Specifically, we would group the existing methods as follows.

**Hidden Neuron Analysis Methods:** The hidden neuron analysis methods decipher the neural networks by interpreting the features that are learnt by constituting neurons individually and collectively [6-11]. For example, Erhan *et al.* [6] proposed to maximize the activation of a single unit to visualize the features learned by hidden neurons. Yosinski *et al.* [9] visualized the activation patterns of neurons when Convnet processes an image or video. Zhou *et al.* [10] proposed to label the hidden units by associating individual hidden units with visual semantic concepts. Cao *et al.* [11] designed a feedforward loop to infer the activation states of hidden layer neurons so as to localize the visual attention to targets. However, the hidden neuron analysis methods only offer qualitative insights, falling short of the optimization of the network structure or performance.

**Model Mimicking Methods:** Model mimicking methods [12-16] find the well-performed models that are more interpretable than existing models. For example, Wu *et al.* [12] utilized tree regularization to regularize a deep model and deliver a directly interpretable decision tree. Fan *et al.* [13] proposed a generalized hamming network based on the fact that neurons calculate generalized hamming distance when bias is appropriately configured. Along with the knowledge distillation methods, Forrest et al. [14] trained a soft tree to mimic the prediction of the original deep models. Albeit seemingly reasonable, the model mimicking methods do not reveal the real mechanism based on which the existing models are successful. In addition, the mimicking models are relatively simple, it is hard to guarantee that these simplified models can imitate the deep neural networks satisfactorily.

**Local Interpretation Methods:** Local interpretation methods are dedicated to understanding the neural networks by perturbing the input and observing the output change of the network [17-21]. Local Interpretable Model-Agnostic Explanations (LIME) [18] uses a binary vector representing the input, in which each bit corresponds to an input feature: one means that the corresponding feature is important, zero means unimportant. Encouraged by the success of LIME, Ross et al. [19] restricted the model to be "right for the right reasons" in the training process with a binary mask to specify relevant features. With robust analysis, training points can be identified that are most responsible for a prediction. P. W. Koh *et al.* [20] explained the prediction of a neural network by identifying most responsible points aided by the influence function in classical robustness statistics. However, the local interpretation methods do not offer direct insight into the inner working of a network and may be interfered by biases of the inputs.

**Mathematical/Physical Models:** Mathematic/physics methods build a top-level connection between deep networks and advanced mathematical or physical theories in order to unlock the mystery of deep learning. Gu *et al.* 2017 [22] offered an elegant explanation of the adversarial mechanism of GAN from the viewpoint of optimal mass transportation. Dong *et al.* 2017 [23] established a correspondence between a deep neural network and ordinary differential equations to guide the design of a network with skip connections. In a perspective on deep imaging [24], it was suggested that "with deep neural networks, depth and width can be combined to efficiently represent functions with a high precision, and perform powerful multi-scale analysis, quite like wavelet analysis but in a nonlinear manner." It was recently shown [25] that convolutional autoencoders are nothing but a framelet representation scheme. In another study, the success of deep learning is attributed to fundamental physical and statistical principles [26]. In spite of being inspiring and in-depth, high-dimensionality and high nonlinearity are aggressively ignored in these explanations, which are actually the key ingredients of deep neural networks.

**Neural Network Visualization:**   Gradient-based methods [58-61] are the mainstream for network visualization. The core idea of such visualization is to alter input images for maximization of the prediction score. Zhou *et al.* [61] found that global average pooling is an attention mechanism of convolutional networks so as to form a localized discriminative network. Practically, network visualization provides a direct understanding to the network dynamics, but it falls short of revealing the cognitive mechanism of the network.

The current artificial neurons were inspired by the functionality of biological neurons. In biological neural system, multiple types of neurons interact for various neural activities. In our earlier studies [27-28], we proposed to construct a neural network with quadratic neurons (in which data are integrated with a quadratic function, instead of an inner product). Pilot results demonstrated that quadratic networks enjoy the merits of a stronger representation ability and a more compact structure than the popular linear neurons. Since the inner product is replaced by the quadratic operation in a quadratic neuron, an individual quadratic neuron can implement not only the XOR logic but also a generalized fuzzy operation. Livni *et al.* [29] proposed to utilize a quadratic activation $\sigma(z) = z^2$ to replace the existing type of activation functions. At an increased computational cost, neurons with quadratic activation are still characterized by linear boundaries, restricting the expressive ability of neurons.

As engineers, we tend to interpret neural networks from the perspective of engineering. A modern computer operates on binary logic, which is metaphorically called an electronic brain. As a complementary metaphor, we consider the deep neural network as an integrated system consisting of fuzzy logic gates. Every fuzzy logic gate has a specific function. When each neuron is viewed as a specific fuzzy logic gate, the overall neural network can be disentangled into these basic units, thereby enabling us to infer how features are extracted and processed by these units, and how information is communicated among these units especially between layers (initially analyzed in this paper) and in loops (to be conducted in the future). In this aspect, quadratic fuzzy logic modules (consisting of quadratic fuzzy logic gates) allow a top-level explanation, such as in light of matrix spectrum theory by which each quadratic module can be topologically delineated and identified by its eigen spectrum. Therefore, we suggest that quadratic neural systems enjoy an interpretability from the fuzzy logic perspective.

In summary, our contributions are of three-folds. First, we propose a fuzzy logic interpretation of quadratic networks. Second, we perform spectral analysis on such a network in terms of the spectral entropy to reveal the relationship between the complexity of the neural network and the properties of the loss function. Third, we investigate the generalization ability of flat minima in the context of spectral analysis.

## II. Generalized Fuzzy Logic Operations

### A. *Generalized Fuzzy Logic*

Aggregation refers to combine multiple values into a single value. Up to now, fuzzy aggregation [30] are extremely rich families of operations. Despite the flexibility in its mathematical expression, fuzzy aggregation is restricted by the range of logic values bounded in [0, 1], which seems unnecessary because data should have different dynamic ranges as appropriate, which can be always normalized if needed. Here we view such an extension as "generalized fuzzy logic". Then, quadratic neurons in artificial neural networks exactly carry on generalized fuzzy logic aggregation operations, each of which is nothing but a fuzzy logic gate.

In our quadratic neurons [27], a quadratic function summates the multiplication of two inner products and one weighted norm term before feeding into a nonlinear activation function. For compact notation, we define $w_{0r} := b_1, w_{0g} := b_2$ and $x_0 = 1$. Then, the output function is expressed as:

$$f(\vec{x}) = \left(\sum_{i=0}^{n} w_{ir}x_i\right)\left(\sum_{i=0}^{n} w_{ig}x_i\right) + \sum_{i=0}^{n} w_{ig}x_i^2$$

$$= \left(\sum_{i=1}^{n} w_{ir}x_i + b_1\right)\left(\sum_{i=1}^{n} w_{ig}x_i + b_2\right) + \sum_{i=1}^{n} w_{ig}x_i^2 + c \qquad (1)$$

The condition $f(\vec{x}) = 0$ forms a decision boundary to separate the input vector space into two regions: the inputs from one region make the function negative, and those from the other region make it positive. The effect of the nonlinear excitation is to produce a fuzzy logic output in a proper range.

Now, let us briefly illustrate the generalized fuzzy logic operations of the conventional and quadratic neurons respectively with the 2D and 3D examples. In Fig. 1, the data cloud with two classes ("red circle" and "green triangle") are split by the boundary shaped by the first-order neurons in Fig. 1(a)-(b) and the quadratic neurons in Fig. 1(c)-(d). We denote two variables in the input vector as A (horizontal axis) and B (vertical axis) respectively. If we assign the logic value "1" to "green triangle" points and "0" to "red circle" points, the neurons in Fig. 1(a)-(c) approximately perform logic operations $\tilde{B}$, $\tilde{A}\tilde{B}$ and $\widetilde{XOR}(A, B)$, respectively. In Fig. 1(d), the aggregation operation of three variables is implemented by a quadratic function $f$, which cannot be approximately described as a conventional logic gate. Conveniently, a generalized logic gate is denoted as a "$f$-logic gate".

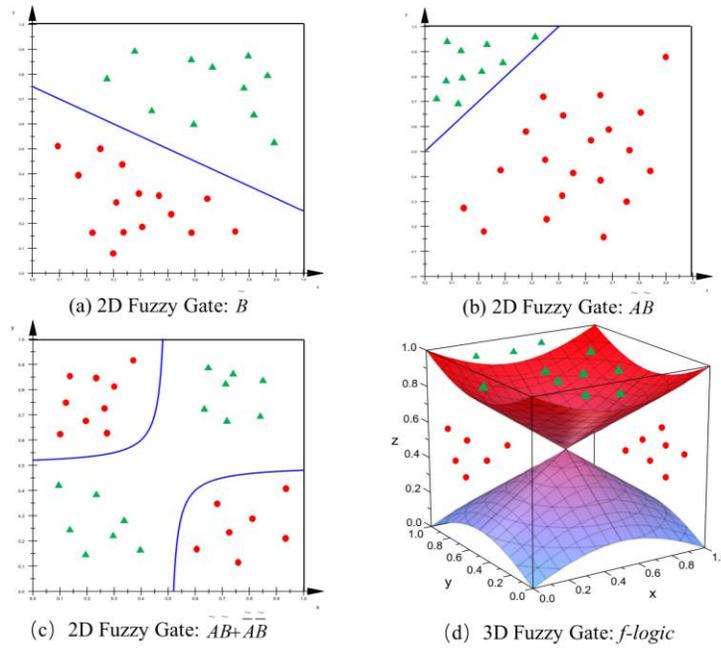

Fig. 1. Generalized fuzzy logic operations implemented using the first-order and quadratic neurons respectively.

## B. *Categorization of Fuzzy Logic Gates by Quadratic Neurons*

For simplicity, the generalized fuzzy operations performed by a quadratic neuron are written as a quadratic logic operation. Accordingly, a quadratic neuron is equivalently called quadratic fuzzy logic gate. To realize a target quadratic function, the quadratic neuron only uses approximate $3n$ parameters for $n$ inputs. In the light of generalized fuzzy logic, every quadratic function implements a unique logic operation. Yet we can still find some high-level topological similarities that are shared by quadratic fuzzy logic gates. For this purpose, we comprehensively characterize quadratic fuzzy logic operations in light of matrix spectrum theory and then categorize them.

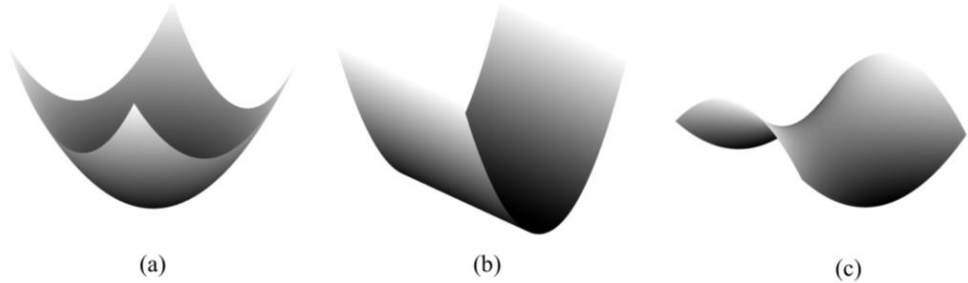

Fig. 2. Topological properties of the quadratic functions in two variables. (a) A parabolic surface, (b) an elongated valley, and (c) a saddle configuration, as characterized by the eigen values and vectors of the corresponding $J$ matrices respectively.

Given the input vector $\vec{x}$ with a dimensionality $n$, its function $f(\vec{x})$ is calculated according to Eq. (1). The parameters $b_1, b_2$ and $c$ control offsets instead of the topological characteristics of the function in a high-dimensional space. Therefore, the decision boundary shape of the function defined in Eq. (1) is essentially the same as that of the following equation:

$$f'(\vec{x}) = \left(\sum_{i=1}^{n} w_{ir} x_i\right)\left(\sum_{i=1}^{n} w_{ig} x_i\right) + \sum_{i=1}^{n} w_{ig} x_i^2 \qquad (2)$$

In linear algebra, the quadratic function can be simplified into a quadratic form. That is, we can write the right-hand side of Eq. (2) as $xJx^T$, where $J$ is an $n$ by $n$ real symmetric matrix with $J_{ii} = w_{ir}w_{ig}$ and $J_{ij} =$ $J_{ji} = (w_{ir}w_{jg} + w_{jr}w_{ig})/2$. The topology of the quadratic function is dominated by the spectrum of $J$, especially by the sign of eigenvalues. Fig. 2 gives us examples to show how the signs of eigenvalues affect the landscape of a quadratic function of two variables. The signs of two eigenvalues are $(+, +)$, $(+, 0)$, $(+, -)$ in Fig. 2(a)-(c) respectively. A "+" indicates that the functional value along the corresponding eigen axis will grow up to positive infinity. A "0" (or very small) eigenvalue suggests a valley along the corresponding axis. The interesting saddle surface is generated in the neighboring axes with opposite signs of eigenvalues.

Since the congruent matrix transformation will not change the index of inertia of $J$, we can categorize the quadratic fuzzy logic operations realized by quadratic neurons according to the positive, negative, and zero indices of inertia of $J$. Considering a quadratic function of $n$ input variables, how many categories should we put the quadratic function into? This is a combinatorial problem that is equivalent to that "for $a + b + c = n$, how many nonnegative, unique integer solutions for $(a, b, c)$." The answer is $(n + 2)(n + 1)/2$, which means that individual fuzzy logic operations can be grouped into $(n + 2)(n + 1)/2$ types. Take a four-inputs quadratic function as an example, there are theoretically 15 semi-topological types of fuzzy logic operations. While in reality, eigenvalues seldom show up as pure zero but a positive or negative number with a relatively small norm compared with other eigenvalues. Hence, we split significant eigenvalues according to their signs, and consider the eigenvalues either positive or negative. Then the combinatorial problem is simplified as "for $a + b = n$, how many nonnegative, unique integer solutions for $(a, b)$." This time, the answer becomes $n + 1$ if we ignore any difference in permutation.

### C. Characterization of Minima

The above-described categorization of quadratic fuzzy logic gates captures the intrinsic shape of the manifold defined by a neuron with an invariability under orthonormal transforms. However, this semi-topological picture is insufficient to describe the operation of the quadratic neuron comprehensively. In other words, although orthonormal transforms have no impact on the index of inertia, they determine the location and orientation of a decision boundary, generally changing the specific function of the neuron. Nevertheless, topologically, neurons in the same class conducts more or less similar fuzzy aggregations, and neurons in different classes behave much more differently. We underline that our eigen-characterization of the quadratic fuzzy gates is semi-quantitative in order to offer a top-level insight. With this motivation, we can develop a spectral analysis method to describe the properties of a neural network, particularly the minima of the network.

For a given well-trained quadratic network with $L$ layers at some minimum, suppose that the number of input variables at the $l^{th}$ layer is $n(l)$, the number of possible types of spectra the neurons in the $l^{th}$ layer may have is $(n(l) + 2)(n(l) + 1)/2$, naturally indexed as $S_i^l$, $i = 1,2, \dots (n(l) + 2)(n(l) + 1)/2$. We represent the total number of elements $S_i^l$ in the $l^{th}$ layer as $N_{S_i^l}$. The whole network can be spectrally characterized as $S = \{N_{S_i^l} | l = 1,2 \dots, L, \ i = 1,2, \dots (n(l) + 2)(n(l) + 1)/2\}$. In the perspective of fuzzy logic, the spectrum $S$ for a given network minimum represents a particular composition of quadratic fuzzy gates implemented by all involved neurons/filters/modules. In a good sense, different minima will embrace different spectra. Therefore, $S$ is the signature/fingerprint/"ID" that helps us to distinguish the different minima of quadratic networks (convolutional or non-convolutional), facilitating evaluation of network-based solutions to a specific problem. It is underlined that spectral analysis is an intrinsic characteristic of quadratic networks and only possible after the introduction of quadratic neurons, since conventional (first-order) neurons are not mendable to matrix spectrum analysis.

### D. Robustness of Spectral Characterization of Minima

The training process for a network usually will not stop at the exact minimum point. Rather, the optimization trajectory will be probably terminated in a neighboring region of some minimum even after a large number of iterations. A potential concern is whether the spectrum $S'$ of neighboring points will be different from the spectrum $S$ of a given minimum or not. More mathematically, the concern is if the eigenvalues of a symmetric matrix are stable with respect to a small perturbation. Luckily, we have the following Hoffman and Wielandt Theorem [31] to guarantee the robustness of our spectral characterization method.

Let $A, B$ are $n$ by $n$ symmetric matrices, $\lambda_i^A$ is the eigenvalue of $A$ and $\lambda_i^B$ is the eigenvalue of $B$, then there exists a permutation $\sigma$ satisfying: $\sum_i^n |\lambda_i^A - \lambda_{\sigma(i)}^B| \leq ||A - B||$, where $||\cdot||$ denotes the Frobenius Norm. This theorem shows that the eigen spectrum of a symmetric matrix is rather robust: as the optimization trajectory comes close to a minimum, the eigenvalues of each neuron will approach the corresponding eigenvalues associated with the minimum. Accordingly, the type of spectra of each neuron will also align with that of the minimum, which makes the spectrum $S'$ of neighboring points consistent with the spectrum $S$.

## III. SPECTRUM DIFFERENCE OF DIFFERENT MINIMA

MNIST is one of the most popular datasets in the machine learning community. The network framework "LeNet-5" [32] achieved an excellent performance on MNIST. To demonstrate the interpretability of quadratic networks by characterizing minima of networks, we built a quadratic neuron based on LeNet to analyze MNIST.

Specifically, we adopted two "convolutional" layers, two pooling layers, one fully-connected layer, and one softmax layer. Here we borrowed the concept of convolution from traditional convolutional neural networks for convenience, although typically kernels in quadratic networks do not represent linear convolutional operations. Nevertheless, these kernels are shifted over an image or pixels as the convolution operations do. We set 32 $3 * 3$ "convolutional" kernels with stride 1 in the first "convolutional" layer, and 20 $4 * 4$ kernels for every channel in the second "convolutional" layer. Both the pooling layers perform down-sampling by a factor of 2 with stride 2. In our quadratic network, all the neurons are quadratic neurons with the activation function ReLU, except for the softmax layer for classification. The network was trained in the TensorFlow environment. All the weights and biases were initialized with a truncated Gaussian function of the standard deviation 0.1. The loss function was defined as the cross-entropy. We set the 55,000 iterations for training, where the first 20,000 iterations were done at a learning rate $10^{-4}$ and then the rate was lowered to $0.4 * 10^{-4}$ for the remaining part of the training process. The results achieved by our quadratic network is in TABLE I, along with the data obtained by other models. For a given well-trained quadratic network, the error rate is the number of instances for which the network gives an incorrect prediction normalized by the number of total testing data. Our quadratic network achieved a lowest error rate 0.76% without data argumentation, which is the state-of-the-art performance on MNIST. Then, we investigated the types of quadratic operations in our quadratic network.

TABLE I: ERROR RATE (%) OF DIFFERENT METHODS ON MNIST

| Methods | Error rate (%) |
|---|---|
| LeNet-5 (no distortion) [62] | 0.95 |
| LeNet-5 (distortion) [62] | 0.85 |
| LeNet-5 (huge distortion) [62] | 0.8 |
| Deep convex net, unsupervised pretrained [63] | 0.83 |
| **Quadratic networks (ours)** | **0.76** |

As shown in Fig. 3, we colored and inflated the neurons in the "convolutional" layers to present our interpretation of this well-trained quadratic network. Due to the dominating role of convolution in the convolutional neural network, we only processed the neurons in the convolutional layers. In the conv1 and conv2 layers, the neurons took 9 and 16 inputs respectively. As we discussed before, the total number of types of quadratic operations are $n + 1$ for $n$-input quadratic neurons. Then, we saw 10 types of quadratic operations for the conv1 layer and 17 types of quadratic operations for the conv2 layer, as shown in Fig. 3. The colors of neurons indicate different types of quadratic operations performed by the neurons. The area of the neuron is proportional to the count of corresponding operations, as labeled inside the neuron. The subsequent squares denote the feature maps that are extracted by the corresponding quadratic operations. We blackened the feature maps from the conv2 layer because the feature maps were exclusively from the solitary neuron.

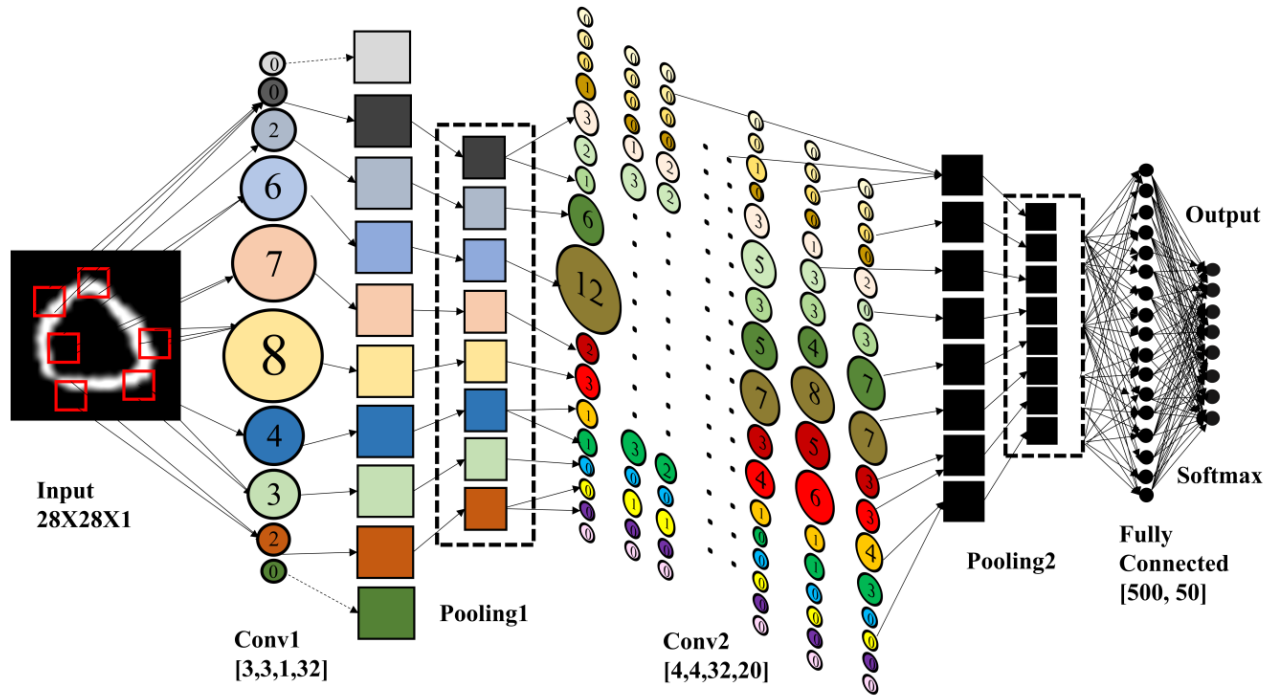

Fig. 3. Quadratic network for recognition of Arabic digits from the MNIST dataset. The neurons in the "convolutional" layers are colored and inflated to indicate the types and frequencies of quadratic operations respectively. For example, the number inside a circle is the number of neurons conducting a particular type of fuzzy operations marked in a unique color.

To evaluate the spectrum associated with a well-behaved minimum, we repeated the training process to have two other minima that achieved the state-of-the-art error rates 0.77% and 0.81% respectively. We counted the number of quadratic operations per each fuzzy gate type for Conv1 and Conv2 layers respectively, and obtained the graphs in Fig. 4. "Type-0" on the horizontal axis indicates the number of positive indices of inertia equal to 0, "Type-1" denotes the number of positive indices of inertia equal to 1, and so on.

As shown in Fig. 4, the spectra of fuzzy logic systems associated with these three minima differ significantly in the first and second convolutional layers respectively. In the Conv1 layer of the three networks, the spectra are irregularly distributed, despite neurons of types 3-5 are dominant in all the cases. For example, the counts of "Types 3-5" in the network with error rate of 0.81% took up 84.38% of all the neurons. In contrast, the spectrum for Conv2 is very close to a normal distribution. Interestingly, the spectra of Conv1 became more and more even and smooth when the error was decreased, which indicates that the involvement of diverse operations contributed to the performance improvement [34].

One point we underscore is that the above spectra offer us a direct understanding of discrepancy of different minima, which is uniquely enjoyed by a quadratic network (both convolutional and non-convolutional) due to its intrinsic quadratic nature. One can argue that similar statistical analysis can be done for conventional networks, such as in terms of the number of positive weights. However, any statistical analysis applicable to conventional networks can also be used in quadratic networks because a conventional network is just a degenerated quadratic network. However, our fuzzy logic interpretation will not be natural to a conventional network because a conventional neuron is handicapped in implementing fuzzy logic operations, such as demonstrated by the famous example XOR.

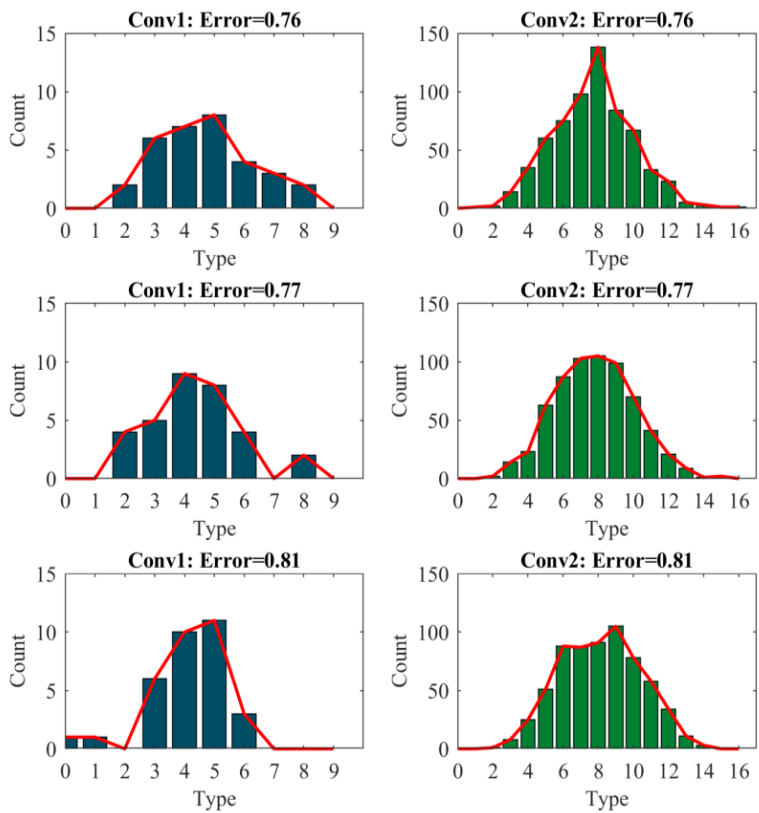

Fig. 4. Spectra of the number of positive indices for three minima respectively.

## IV.  SPECTRAL ANALYSIS AND CATEGORIZATION

In the preceding section, we have scrutinized the spectra of three different minima in a well-trained quadratic network for recognition of Arabic digits in MNIST, which suggests a spectral analysis method to decipher the neural network for interpretability. In this section, several applications of spectral analysis are presented: we first study the loss change in terms of good minima as the network goes wider. Then, we apply spectral analysis to test if it is correct that flat minima generalize well, and propose a new measure to evaluate the network structure.

The experimental details are follows: the experiment was based Tai Ji benchmark, one was generated in reference to the most popular pattern Tai Ji, which represents, to many people, the essence of Chinese culture. The training dataset was made by gridding the two vectors -1:0.05:1 and -1:0.05:1 and then labeling the points with "+" and "o" according to the Tai Ji pattern. Finally, the 1,245 instances were synthesized as the training dataset, as shown in Fig. 5. To our best knowledge, the Tai Ji benchmark was not used before.  Three different network configurations 2-6-6-1, 2-8-6-1 and 2-10-6-1 were selected to represent gradually widened network. For a specific network, we conducted training using the standard backpropagation algorithm after randomly initializing weights and biases. The learning rate was set to 0.004. The number of iterations was set to 1,000.

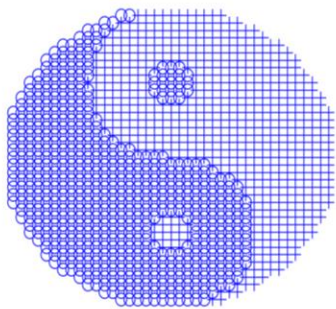

Fig. 5. Tai Ji benchmark used in this study, which is not used in the machine learning community before.

### A. Network Complexity and Loss Function

For a given training dataset, the optimization landscape of the loss function with respect to parameters of a network is fundamentally controlled by the network structure. At the same time, this landscape is also strongly influenced by the form of the loss function; for example, the Euclidean distance versus the information divergence. The non-convexity of the loss function determines the easiness in finding the minimum with a low generalization error [38]. It is of significance to investigate how the network structure affects the landscape of the loss function. However, due to the dimensionality curse, it is prohibitive to study the loss function directly by looping through all variable values in variable space. Hence, current studies were either theoretical [33-37] or visual [38-41]. For instance, by aggressively assuming neurons were linear, Hardt & Ma [41] showed that local minimum was the global minimum by the principle of identity parameterization. Visualization methods such as 1D linear interpolation [39-40] and 2D contour plot [41] were widely used to study the "sharpness" of minima and the trajectories of optimization methods.

For the quadratic network amendable for fuzzy logic interpretation, we can now characterize a minimum by its spectrum $S$, such as we did in Fig. 4, where we plotted the spectra of a convolutional layer for three minima respectively. We are especially interested in the spectra of minima that can achieve superior performance on the training dataset. Based on our visual checking, the predicted patterns from networks can by-and-large keep the Tai Ji shape if the classification performance is over $\frac{1200}{1245} = 96.39\%$. Hence, we refer those minima that have a classification performance over 96.39% on the training dataset as "good minima". Please note that good minima may not be truly "good" because they may overfit when the network is generalized to new data, but identifying the "good" minima is the first step to understand the loss function. Accordingly, we also define the minima with performance lower than 96.39% as inferior minima. We will look into the generalization ability in the next subsection. As we know, the important questions for the landscape of the loss function are how many good minima in total and how sharp each of these minima is. As we argued, the spectrum $S$ of the whole network works somewhat like the "ID" for the minimum of the network, thereby allowing us to differentiate different minima on a high level. Since the spectrum contains critical information on the minimum, how many good minima are there of the loss function is directly related to how many types of spectra $S$ are favorable. Furthermore, acquiring a sharp minimum requires to specify the initial value with high precision. In contrast, a flat minimum merely requires a rough initial value. In other words, with random initialization, the sharpness of a minimum is closely correlated with the probability of getting this minimum. Said differently, the sharper the minimum is, the harder to find this minimum. Consequently, the sharpness of a minimum determines the frequency of getting the spectrum of that minimum. Aside from discussing the minima of the loss function, we can also evaluate the complexity of the loss function in terms of good minima through spectral analysis.

• *Experimental results on Tai Ji dataset*

When the optimization process was trapped at minima, we calculated the performance over the dataset. We recorded the spectral types for the first hidden layer for good minima and dropped bad points. In the fuzzy operation view, the neurons with two input variables in the first layer must belong to one of three quadratic fuzzy operation types based on their matrix eigenvalues, denoted by $(+, +), (+, -), (-, -)$. With respect to the first hidden layer neurons in the 2-6-6-1 network configuration, if one minimum has 2 neurons perform the $(+, +)$ operations, 3 neurons conduct the $(+, -)$ operation, and 1 neuron does the $(-, -)$ operation, then the spectral type of this minimum is compactly denoted as $(2,3,1)$. The spectral type in the second hidden layer is not recorded because herein the neurons in the three structures have different numbers of inputs, which cannot be compared directly in this feasibility study (certainly, they can be compared after certain normalized in a follow-up study). We repeated training and recording until enough good minimum points were recorded. It can be proven that given the number of neurons $n$ there are potentially $C_{n+2}^2$ types of spectra. Therefore, for the configurations 2-6-6-1, 2-8-6-1 and 2-10-6-1 with 6, 8, 10 neurons respectively in the first hidden layer, there are 28, 45, 66 types of spectra at maximum that can potentially show up for good minima. Hence, we respectively recorded 56, 90, 132 good minima for structures 2-6-6-1, 2-8-6-1 and 2-10-6-1 to keep the ratio of good minima identical for the configurations. The results of the spectra and corresponding frequencies are shown in Fig. 6.

Our experiment retrieved 16, 24, 37 types of "good" spectra for the 2-6-6-1, 2-8-6-1 and 2-10-6-1 configurations respectively. The increased types of good spectra from 2-6-6-1 to 2-10-6-1 definitely means the increased numbers of good minima for the corresponding loss functions. This result is not surprising because the newly added neurons, coupled with old neurons, in the first layer should create new minima. These new minima are from effective combinations of fuzzy logic modules.

Furthermore, we can visualize the frequency of spectral components in green fonts in Fig. 6. For clarity, we only marked some main spectral components and their corresponding frequencies. The frequency of a spectral component does matter because it reflects the sharpness of a good minima. In the loss function of the network 2-6-6-1, the minima with spectrum type $(2,2,2)$ are dominating with a frequency 17.86%, while the dominant minima in the loss functions of the networks 2-8-6-1 and 2-10-6-1 account for 14.44% and 9.09% respectively, which means the optimizer are easier to converge to the dominant minima in 2-6-6-1 compared to that in 2-8-6-1 and 2-10-6-1 configurations. Additionally, we calculated the variance of frequency of spectral components for the three configurations, which are 0.0017, 0.0011, 0.0004 for 2-6-6-1, 2-8-6-1 and 2-10-6-1 respectively. These data indicate that the sharpness measures of different minima of the loss function of 2-10-6-1 are closer with each other than that of 2-8-6-1 and 2-10-6-1.

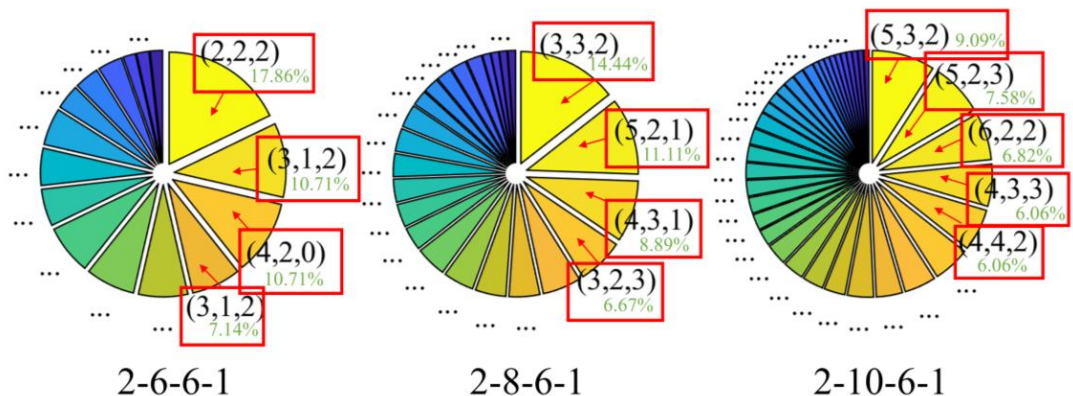

Fig. 6. Spectral analysis on good minima. The number of spectral types is correlated to the number of good minima. The frequency of a spectrum suggests the flatness of a good minimum.

Generally speaking, the more minima the loss function contains, the less organized the minima, and the higher the complexity of the loss function. To extract an overall characteristics of a given network, we define the entropy of good minima (EGM) as $-\sum_i p_i \lg_2(p_i)$, where $p_i$ is the probability of getting the $i^{th}$ good minimum. This measure describes the complexity of the optimization landscape defined by the loss function specific to the network, and also reflects the complexity of the loss function. In reality, we do not have a method to characterize good minima individually, but we can approximately calculate EGM by replacing spectral components with minima and then probabilities of minima with frequencies of spectral components. It turns out that the entropy of the loss functions of 2-6-6-1, 2-8-6-1 and 2-10-6-1 are 3.7269, 4.2366, 4.8233 respectively. As expected, the added neurons indeed increased the complexity of the loss function. Generally, structural complexity means functional complexity, being consistent to the principle in biology that structure determines functionality.

- *Theoretical Analysis on EGM*

Indeed, EGM will ascend as the network goes wider. Without loss of generality, it is assumed that the network is a multilayer perceptron (MLPs), and our conclusion also holds for convolutional models after some adaptions. By wider networks we mean that the number of neurons in every hidden layer is at least the same as that of the reference network; for example, the structure 2-5-3-2 is not wider than structure 2-3-4-2.

**Proposition:** Entropy of Good Minima (EGM) of the MLP will not decrease if this MLP is augmented with new neurons.

**Proof:** By definition, the loss function of the parent network has at least one good minimum; otherwise this network is not well structured to compute EGM. Without loss of generality, it is further assumed that the loss function of a network $\phi$ has $k$ good minima. If the chance of finding the $i^{th}$ minima is $p_i^{\phi}$, the corresponding EGM is $\sum_i^k p_i^{\phi} \lg_2(p_i^{\phi})$. Next, we add new neurons into $\phi$, thereby building a wider network $\Phi$.

Then, we target good minima of the network $\Phi$ formed by setting parameters inside the newly added neurons to zeros. At the same time, considering that the added neurons and old neurons are of the same structure, we can exchange new neurons with the same number of old neurons in the same layer, and set their parameters to zero. Clearly, the total number of good minima will be enlarged to $Ck$, where $C$ denotes a certain combinatorial constant. Due to the equivalence, the probabilities of finding these $Ck$ minima end up being divided by $C$, as $\frac{p_1^{\phi}}{C}, \dots, \frac{p_1^{\phi}}{C}, \dots, \frac{p_k^{\phi}}{C}, \dots, \frac{p_k^{\phi}}{C}$ respectively. Then, the EGM of $\Phi$ is computed as follows:

$$-\sum_i^k C\left(\frac{p_i^{\phi}}{C}\right)\log\left(\frac{p_i^{\phi}}{C}\right) = -\sum_i^k p_i^{\phi}\log\left(\frac{p_i^{\phi}}{C}\right)$$

$$> -\sum_i^k p_i^{\phi}\log(p_i^{\phi}) \tag{3}$$

However, the added neurons can work nontrivially, which means that the added neurons together with old neurons can create new good minima in addition to the just-mentioned $Ck$ good minima. Let us say that there are $l$ new good minima so created and the corresponding probabilities of finding them are $p_1^{\phi}, \dots, p_l^{\phi}$ respectively. Then, the new minima will take away the likelihood of the previous minima by a factor $\alpha$, $\alpha = 1 - \sum_i^l p_i^{\phi}$, and the EGM of the new network is

$$-\sum_i^l p_i^{\phi}\lg_2(p_i^{\phi}) - \sum_j^k C(\alpha p_j^{\phi}/C)\lg_2(\alpha p_j^{\phi}/C) \tag{4}$$

Due to the monotonically decreasing property of $-x\lg_2 x$ over (0,1), the right term of Eq. (4) satisfies:

$-\sum_j^k \left(\frac{\alpha p_j^{\phi}}{C}\right)\lg_2\left(\frac{\alpha p_j^{\phi}}{C}\right) > -\sum_j^k \left(\frac{p_j^{\phi}}{C}\right)\lg_2\left(\frac{p_j^{\phi}}{C}\right)$. Since $-\sum_i^l p_i^{\phi}\lg_2(p_i^{\phi}) > 0$, we have

$$-\sum_{i}^{l} p_i^{\phi} \lg_2(p_i^{\phi}) - \sum_{j}^{k} C\left(\frac{\alpha p_j^{\phi}}{C}\right) \lg_2\left(\frac{\alpha p_j^{\phi}}{C}\right)$$

$$> -\sum_{j}^{k} p_j^{\phi} \lg_2\left(\frac{p_j^{\phi}}{C}\right) > -\sum_{j}^{k} p_j^{\phi} \lg_2\left(p_j^{\phi}\right) \tag{5}$$

By Eqs. (3) and (5), we finish our proof.

EGM describes the complexity of the loss function of a network. As the above proof shows, if the network goes wider, the complexity of the loss function will increase. It seems that structural enhancement will result in the functional enhancement. Can this trend hold when network goes deeper? The answer is yes, and a very similar proof can be given, which is not included here for conciseness.

In reality, how to find every individual good minimum is still far from being clear, a practical training strategy is needed. For quadratic networks, we can use spectral analysis to assess good minima, thereby labeling/ranking the good minima into several classes. Hence, spectral signature seems a helpful hint on good or bad minima. In our aforementioned experiment, the entropy values for the 2-6-6-1, 2-8-6-1 and 2-10-6-1 configurations are 3.7269, 4.2366, and 4.8233 respectively. This result agrees with the non-decreasing trend of EGM, predicted by our theoretical proof.

### B. Flat Minima and Network Evaluation.

Now, we extend the previous results to the generalization ability of good minima and further discuss whether flat minima would generalize better and how to design the network structure. Next, we shed light on how to evaluate the network via spectral analysis.

• *Generalization Ability of Flat Minima*

As we brought up earlier, we need to initialize network parameters with high precision in order to obtain sharp minima instead of being trapped elsewhere. In contrast, flat minima require loose specification of parameters. In the terminology of information theory, fewer information is needed or we have better chance to pick up flat minima [42]. In general, it is claimed that less description length means low network complexity and high generalization performance [42], as indicated by Occam's razor principle. Hence, most people believe that flat minima can generalize better than sharp minima. Hinton and Van Camp [43] justified the generalization ability of flat minima by the Bayes argument via KL-divergence. However, there are papers arguing that sharp minima could generalize better for deep models [44]. Here, with spectral analysis, we can compare the generalization performance of flat and sharp minima in the case of quadratic networks.

In the forerunner experiments, we recorded 56, 90, 132 good minima and corresponding spectra respectively for the three configurations. Here we first constructed 7,825 instances as a test dataset by gridding two finer vector -1:0.02:1, then we evaluated the generalization performance of these good minima. Taking the 2-6-6-1 network as an example, green line segments connecting green triangles in Fig. 7 show the mean generalization performance of every spectrum of the 2-6-6-1 structure, while bar shows the frequency number of obtaining the corresponding spectrum.

As we stated, the sharpness of a minimum is more or less reflected by the frequency of getting this minimum. Therefore, minima of spectrum (3,0,3) is one of the sharpest minima. However, it has the best mean generalization ability admitting the performance of 98.02%. This is a counterexample demonstrating that sharp minimum can generalize well. On the other hand, aside from showing a solitary counterexample, we also need to conduct statistical comparison, since the argument that flat minima generalize better is the general opinion. Hence, we compared the mean generalization performance of spectra with top 7 highest frequencies to that of remained spectra. Because top 7 spectra accounts for 67.86% frequencies of minima, average flatness of these 7 spectra should be larger than that of the remained spectra. In this way, a statistical sense is made. It turns out that the generalization performance of top 7 spectra is 97.17%, while the generalization performance of the remained spectra is 97.48% for the 2-6-6-1 network. Similar calculations for the spectra of the 2-8-6-1 and 2-10-6-1 networks were done as well. In the 2-8-6-1 configuration, 24 different types of spectra showed up, the mean generalization performance of top 5 spectra (accounting for 41% frequency in total) is 97.49%, which is slightly lower than 97.50% of the remained spectra. Consistently, in the 2-10-6-1 configuration, 37 spectra were identified, and the performance of top 5 spectra (accounting for 32.58% frequency in total) is 97.58%, lower than 97.70% of remained 32 spectra.

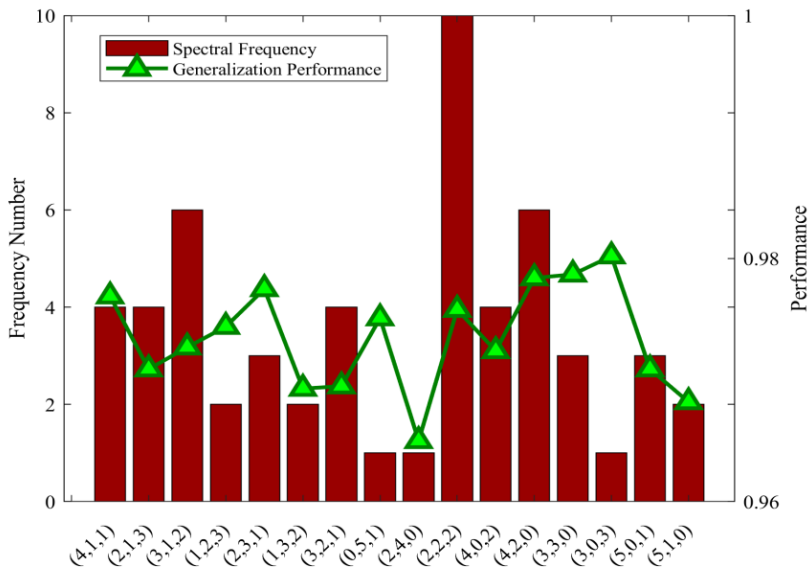

Fig. 7. In the 2-6-6-1 network, the sharp minima generalize better over the Tai Ji dataset.

Our data demonstrate that sharp minima may generalize better than flat minima on average. Since there are other experiments showing that flat minima generalize better, we argue that there may not be necessarily a simple relationship between the sharpness of minima and the generalized ability.

• *Network Evaluation*

Recently, more and more researchers are exploring guidelines for design of network architectures [45-54]. In [47-48], the target network was evaluated according to their performance on a given cross-validation dataset using a reinforcement learning technique. In [45-46], sparsity regularization was employed to generate a compact network that admits performance better than that of a large network. In [54], an attempt was made to widen the network instead of making it deeper. How to assess the structure of a given network and optimize its structure and/or parametrization is still an open question, mainly due to the black-box nature of neural networks. Here, we propose a novel measure to evaluate the network based on spectral analysis.

For deep learning engineers, they not only like network candidates that have minima with a strong generalization ability but also prefer those candidates from which they can obtain desirable minima more easily in the training process. Said differently, engineers hope that the minimum will be obtained easily and yet possess good generalization ability so that they can finish tuning parameters with least time to meet the performance requirement. Based on this consideration, we define a new measure as: $M = \sum_i p_i K_i$, where $p_i$ is the probability of the $i^{th}$ good minimum and $K_i$ is corresponding generalization performance. The network with high $M$ value will enjoy a good balance between trainability and performance. To calculate $M$, we can again replace probability with frequency of spectral components as an approximation.

Next, we use $M$ to evaluate the three network structures over Tai Ji data in terms of spectral frequency and generalization performance. The score of the three structures 2-6-6-1, 2-8-6-1 and 2-10-6-1 are 0.9741, 0.9741 and 0.9762 respectively. Since 2-10-6-1 achieved the highest score, we believe that the network structure 2-10-6-1 is the best among these structures. As for 2-6-6-1 and 2-8-6-1, their scores are identical, we believe the difference between them is slim in trainability.

## V. ENTROPY VIEW OF TRAINING

The network training usually starts with randomly initialized parameters. After hundreds and thousands of iterations, the gradient search algorithm will often converge to a certain stable point. With spectral analysis, we have a top-level understanding of the training process. We built a quadratic residual network by replacing all the filters of ResNet-20 [64] with "quadratic" counterparts, referred as quadratic-ResNet-20, and then trained it on the CIFAR-10 benchmark. The input data to ResNet-20 are $32 \times 32$ images. The first layer performs $3 \times 3$ convolutions into 16 feature maps, is followed by 9 residual blocks (18 layers) using $3 \times 3$ convolutions to generate the 32,16 and 8 feature maps as three stages respectively. Each stage has three residual blocks and six convolutional layers. The final layer is a fully-connected layer. All the training hyper-parameters of the quadratic-ResNet-20 network are the same as that in the configuration for training the original ResNet-20 except for the initialization of filters and biases. For all the layers, we initialized $w_r$ using the Xavier initialization method, $w_g$ and $w_b$ were set to 0. Biases $b_r$ and $b_g$ were initialized to 0, 1 respectively. In this way, quadratic terms $(w_r x + b_r)(w_g x + b_g) + w_g x^2$ collapsed into linear terms $(w_r x + b_r)$. The reason behind this initialization is that the quadratic terms should be learned instead of pre-determined. The training process comprises 80,000 steps.

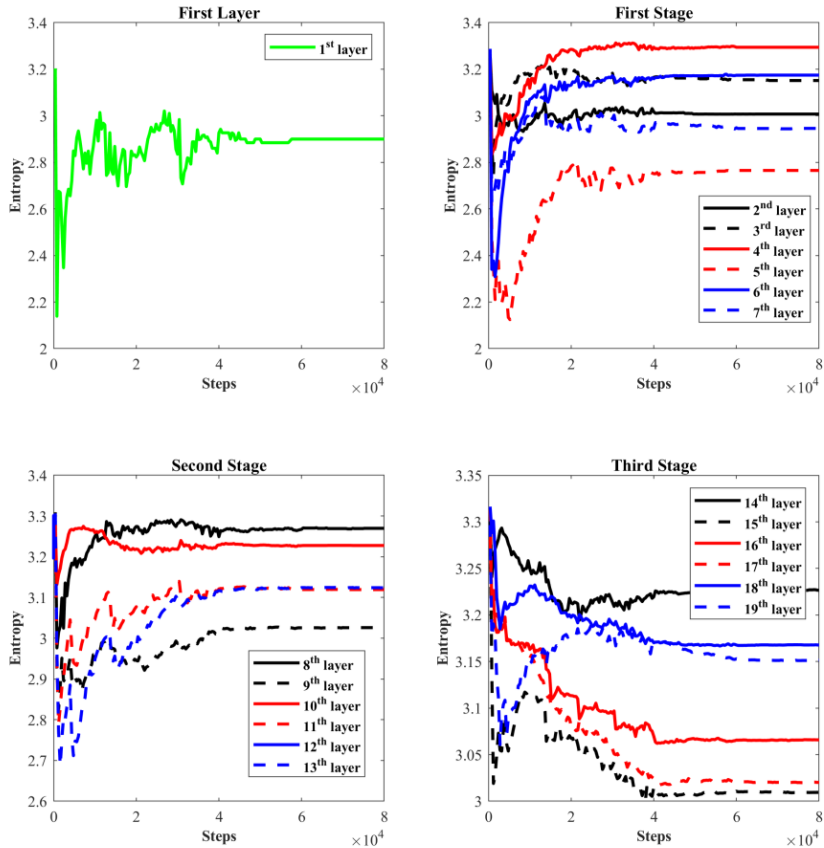

Fig. 8. Dynamics of entropy for convolutional layers of quadratic ResNet-20 on the CIFAR-10 benchmark. The curves in the same color belong to the identical residual block. Through the whole training process of 80,000 steps in total, the entropies of each convolutional layers are recorded in every 400 steps.

Resembling to what we did before, we applied our spectral analysis method to the feature maps of each layer in every 400 steps through the training process. For that purpose, the entropy $-\sum_i p_i \lg_2 p_i$ was computed, where $p_i = \frac{Number\ of\ feature\ maps\ with\ the\ i^{th}\ spectrum}{The\ Number\ of\ feature\ maps}$. The entropy measures how organized feature maps are, as widely used in statistical physics. Since the number of training steps is 80,000, we recorded 200 instances in total.

Figure 8 shows the dynamics of entropy for convolutional layers. In each subfigure, the curves in the same color are from two subsequent convolutional layers in a given residual block. As shown in Figure 8, the entropy values of the layers in the first two stages go up while the counterparts in the third stage go down. The trends of entropy values reflect the converging behavior of the neural network. In the early training stage (before 120 steps), the entropies changed radically, after that the entropies remain status quo, which suggests the convergence of the training process. One note we highlight is the robustness of our spectral characterization: in the late training phase, the entropies become quite stable without significant oscillations, which agree with the theoretical analysis. Interestingly, in each residual block except for the first block, an earlier layer has a higher entropy average than a later layer. This implies that filters in an earlier layer are less organized and tend to learn diversified primitive features.

## VI. DISCUSSIONS AND CONCLUSION

With the concept of "generalized fuzzy operations", the quadratic neurons are categorized into various eigen types, which can be used to implement various fuzzy operation gates. Categorization is the first step to figure out the role of quadratic neurons in a neural network, thereby allowing a fuzzy logic understanding of the functionalities of neurons, layers, and networks. As a positive outcome, our spectral analysis method has given valuable hints on the functionalities of the convolutional layers, minima of the loss function as the network goes wider, network design, and the training process. More efforts are needed to unveil the deep mechanisms that govern the neural network of quadratic neurons, such as the effect when a network goes deep, wide, both ways, or connected with in graphs and with memories.

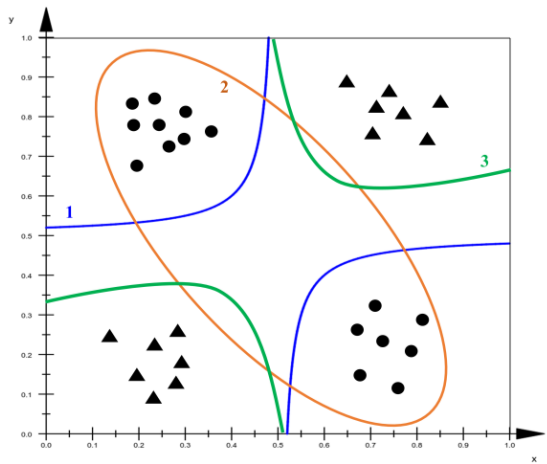

Fig. 9. Eigenvalue characterization fail to differentiate decision boundary curves 1-3.

Eigenvalue characterization is a feasible path to describe the functions of neurons and networks, which is well fitted to quadratic networks. In our fuzzy interpretation, each spectral type represents a generic family of functional modules embedded in quadratic neurons. As we asserted before, our eigenvalue characterization is based on the intrinsic semi-topological feature of quadratic neurons. However, the decision boundary formed by a neuron must be addressed when we move our spectral analysis from top to bottom levels [55]. This is the reason why our spectral analysis is only a global approximation to the true picture. For example, with the use of one quadratic neuron to classify an XOR-pattern in Fig. 9, all three different decision boundary curves implementable with a quadratic neuron, denoted in three colors respectively, are qualified to finish this task.

Our eigenvalue characterization method can tell curve 2 from curves 1 and 3 but fail to differentiate curves 1 and 3, although curves 1 and 3 are two different solutions for XOR-like task. Despite this limitation of spectral analysis, we can still get reliable results in a statistical sense. In contrast, the networks with conventional neurons do not offer such a unique yet convenient angle to characterize the network.

Through interconnected fuzzy gates, complicated system tasks are accomplished by composite functions whose building blocks are fuzzy operations or quadratic neurons. Interpreting the inner working of a neural network in terms of basic fuzzy logic operations is a typical case of reductionism. It is feasible to understand how a complex system works by figuring out the functions of components of that system and their interconnections. In the history of engineering, modular methods are often effective, such as for software development [56] and integrated circuits design [57]. By constructing neural networks in this modular way, there are opportunities to optimize the network architecture design and improve the end-to-end performance.

In Boolean algebra, logic expressions can be possibly simplified based on the truth table. The simplified expression is logically equivalent to the expression before simplification. Our hypothesis is that the generalized fuzzy logic expressions are reducible as well. Then, a complicated neural network consisting of neurons representing generalized fuzzy operations can be made equivalent to more compact neural networks. As the next step, we plan to find the equivalent networks of a well-trained network so that our fuzzy logic approach offers not only interpretability but also guidelines for deep learning practice.

In conclusion, with specific examples we have demonstrated that a fuzzy logic interpretation can be given to any neural network. Aided by spectral analysis and entropy measure, unique insights can be obtained in the context of machine learning with neural networks. Our approach is complementary to other efforts offering interpretability of neural networks, and whose potential is yet to be further explored.